# YouTube Ad View Sentiment Analysis using Deep Learning and Machine Learning

Tanvi Mehta
Student at Department of Computer Engineering
Pimpri Chinchwad College of Engineering
Pune, India

Ganesh Deshmukh
Prof. at Department of Computer Engineering
Pimpri Chinchwad College of Engineering
Pune, India

## ABSTRACT
Sentiment Analysis is currently a vital area of research. With the advancement in the use of the internet, the creation of social media, websites, blogs, opinions, ratings, etc. has increased rapidly. People express their feedback and emotions on social media posts in the form of likes, dislikes, comments, etc. The rapid growth in the volume of viewer-generated or user-generated data or content on YouTube has led to an increase in YouTube sentiment analysis. Due to this, analyzing the public reactions has become an essential need for information extraction and data visualization in the technical domain. This research predicts YouTube Ad view sentiments using Deep Learning and Machine Learning algorithms like Linear Regression (LR), Support Vector Machine (SVM), Decision Tree (DT), Random Forest (RF), and Artificial Neural Network (ANN). Finally, a comparative analysis is done based on experimental results acquired from different models.

## 1. INTRODUCTION
With the rapid growth of Online Social Networks (OSNs), communication platforms have become popular among the public. This has led an excess population to share, search and interchange information and data with each other without any concern about geographical distance. The volume of data created through Online Social Networks (OSNs), especially YouTube, is massive [1]. YouTube, a social platform that allows free sharing of videos which makes online watching of videos easy. People can create and even upload videos on their own. It has become one of the most popular websites, with viewers watching about 6 billion hours of videos monthly.

The number of YouTube users escalates each day and they tend to post each and every detail or information regarding the variety of topics and trends they are concerned about. This lets all the viewers put forward their opinions, likes, dislikes, etc. on these platforms. This leads to the creation of a large amount of data that has to be analyzed [2]. Here, the importance of sentiment analysis establishes. Thus, this information or data visualized through technical domains like Deep Learning and Machine Learning is vital for people creating advertisements and posting them on YouTube.

The content or video creators are being paid by YouTube advertisers on the basis of ad views, clicks, shares, etc. for the products, goods, and services being marketed [3]. The aim of advertisers behind doing this is to estimate the ad views based on other criteria and metrics like comments, dislikes, likes, etc. Analyzing this information manually is a very tedious task. This may be time-consuming and even the results won't be accurate and efficient. This will therefore affect wrong predictions of data leading to the declination of profit for products. Currently, automation in technology is the most demanded tool [4]. Many Deep Learning and Machine Learning techniques are suited for data visualization.

To the best acquaintance of this study, there is less exploration and experimentation done on YouTube sentiment analysis. Therefore, this paper proposes the prediction of YouTube ad views using technological advancements. Techniques like Linear Regression (LR), Support Vector Machine (SVM), Decision Tree (DT), Random Forest (RF), and Artificial Neural Network (ANN) are used and based on their results comparative analysis is done. In section 2, the past research work done is summarized. Section 3 explains the framework of the proposed methodology. In section 4, the techniques used in the proposed research are briefly described. Section 5 illustrates in detail the implementation flow of the proposed work. The analytical outcomes and evaluation are tabulated in section 6. The conclusions and future scope are discussed in section 7.

## 2. LITERATURE SURVEY
A lot of research on sentiment analysis is done. Data or information generated from social media platforms like Instagram, Facebook, and Twitter has been analyzed. This section explains, in brief, about the work done on this topic. Different works proposed are described here:

Rudy Prabowo et al. in 2017, [5] proposed a combined rule-based classification of machine learning models. It tested the reviews of products, movies, and comments of MySpace. The outcome showed that a hybrid categorization can revamp the effectiveness of the classification. They also proposed a complementary, semi-automatic approach in which all the classifiers can acquire a good level of efficiency.

Van Cuong Tran et al. in 2018, [6] explored the new approach that is grounded on a feature ensemble model comprising of fuzzy sentiment associated with tweets by considering lexical, position, etc. words by polarity. They experimented on real data that resulted in the effectiveness of the performance of algorithms.

Doaa Mohey et al. in 2019, [7] surveyed the challenges faced by sentiment analysis based on the relevant techniques and approaches. The paper summarizes the challenges with respect to the review structure. They distinguished challenges into two types and focused on the factors affecting challenges.

Ashima Yadav et al. in 2020, [8] explored the taxonomy of sentiment analysis and surveyed the deep learning models that are explicitly used in it. They also did a comparative analysis of the datasets that are popular and key features of it with the accuracies acquired from various models.

Nhan Cach Dang et al. in 2020, [9] reviewed the advancements in deep learning models which could be the solution for sentiment analysis exceptions. They applied the word embedding and Term Frequency-Inverse Document Frequency (TF-IDF) to the datasets and concluded with a comparative study of results obtained through experiments on the input features and different models.

## 3. METHODOLOGY

This section delineates the usual process of implementing sentiment analysis. The schematic representation of the proposed approach is demonstrated in fig. 1. Brief discussion of the flow is done below:

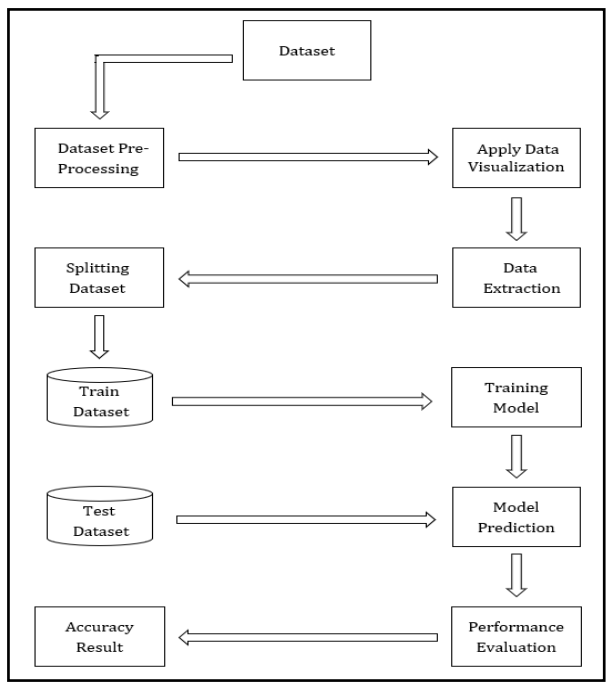

**Fig. 1: Schematic Representation of Methodology**

### 3.1 Dataset
The dataset includes the metrics that have to be cleaned and visualized.

### 3.2 Dataset Pre-processing
It is a technique where data has to be cleaned and manipulated to get and extract the significant data from it. The extracted data is further visualized using the data visualization techniques.

### 3.3 Splitting Dataset
In this step, the dataset that is required gets bifurcated into training and testing sets. Normally, the training – testing proportion is 80% - 20% respectively.

### 3.4 Training Model
This step comprises the implementation of a classification model with help of classifiers like LR, SVM, DT, etc.

### 3.5 Performance Evaluation
All the classifiers or models that are implemented, achieve some or the other accuracies. The classifier obtaining higher accuracy will be the most efficient one. In the case of Root Mean Squared Error (RMSE), the classifiers having the least error score is the more accurate one.

## 4. TECHNIQUES
The proposed techniques for YouTube Ad view Sentiment Analysis are briefly explained in this section. The techniques are Linear Regression (LR), Support Vector Machine (SVM), Decision Tree (DT), Random Forest (RF), and Artificial Neural Network (ANN).

### 4.1 Linear Regression
In Regression, a graph is being plotted between the variables which appropriately suit the data points. It depicts a curve or line which passes through every data point on this wise that it shows the minimum vertical distance between the regression line and the data points. Linear Regression is the supervised algorithm, that is simple and based on the statistical regression method. It is generally applied to problems dealing with predictive analysis. It shows the relationship between the independent variable (X-axis) and dependent variable (Y-axis), as shown in fig. 2. The linear regression model depicts the slope giving a straight line describing the relationship within the variables.

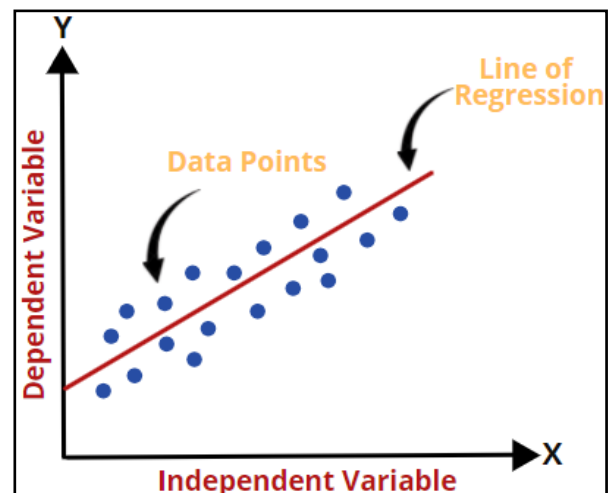

**Fig. 2: Linear Regression Technique**

### 4.2 Support Vector Machine
SVM is not only applied for classification, but also regression challenges. In this supervised algorithm, the data is segregated using the best decision boundary, i.e., the hyperplane. It selects the extreme points called as support vectors, that help in constructing the hyperplane. It also has the positive hyperplane which passes through one (or more) of the nearest positive points and a negative hyperplane which passes through one (or more) of the nearest negative points. The optimal hyperplane is the one where the margin, distance between positive and negative hyperplane, is maximum as shown in fig. 3 [10].

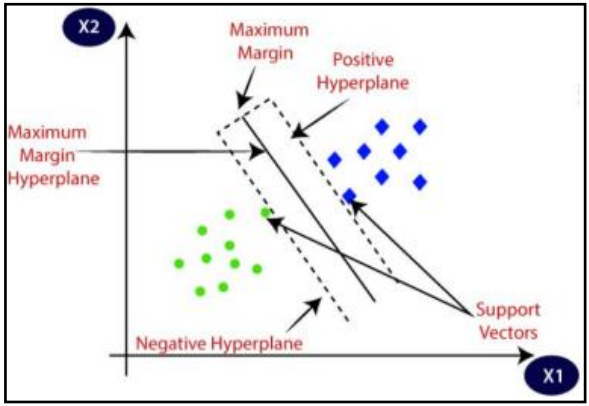

**Fig. 3: Support Vector Machine Technique**

## 4.3 Decision Tree
The decision tree is not only applied for classification, but also for regression challenges. As the name suggests, this supervised algorithm uses the tree structure to depict the predictions that result from a series of element-based splits. It begins with the root node and ends with a decision made by leaves as depicted in fig. 4 [11]. It acquires some basic terminologies like root nodes, leaf nodes, decision nodes, pruning, sub-tree, etc. This technique possesses a bunch of if-else statements where, it checks if the condition is true, if yes, then it moves forward to the next node attached to that decision.

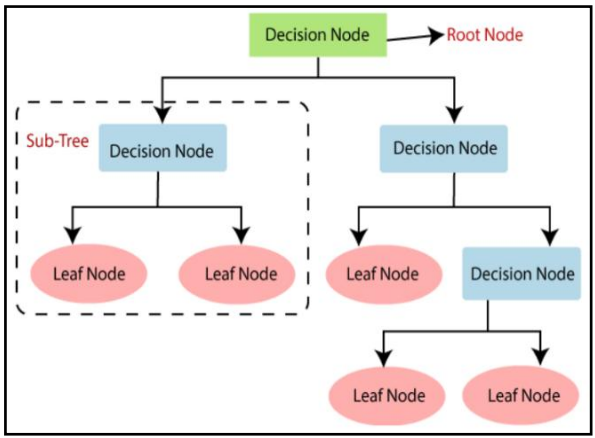

**Fig. 4: Linear Regression Technique**

## 4.4 Random Forest
The Random Forest, being a supervised algorithm, is based on ensemble learning. Here, different classifiers are incorporated to deal with a problem to enhance the efficiency of the model. It contains several decision trees on different subsets of the same dataset and the average is taken for improved accuracy as shown in fig. 5 [12]. Multiple decision trees predict the output and the random forest takes the majority of these predictions for the final output. As eventually the number of individual trees in the forest increase, the accuracy also increases. It is assumed that in a random forest, all the trees predict the correct output. In the first phase, the random forest is created by combining different trees. In the second phase, the output is predicted for all the decision trees. Random forest is a very efficient way of solving classification and regression problems in machine learning.

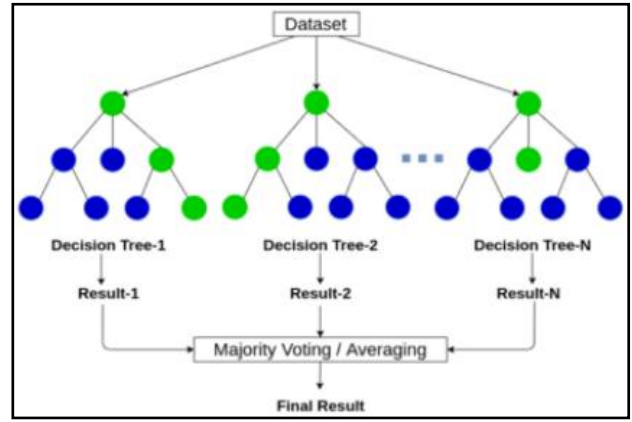

**Fig. 5: Decision Tree Technique**

## 4.5 Artificial Neural Network
ANN is a more advanced technique than other primitive methods. Initially, the information or data is sent to the input layer and then is transmitted to the hidden layers. At the initial point, the interconnection is established between two of these layers, and weights are assigned randomly to every part of the input. Then, each input neuron is attached to the bias, and, the activation function processes that and a combination of weights. The responsibility of deciding the node to fire towards feature extraction, is of the activation function. Then, the output is computed. This entire process is called as Forward Propagation. After acquiring the final output, the comparison is done and error is studied. Furthermore, to reduce the error, the weights are upgraded to backward propagation as illustrated in fig. 6. This is a continuous process until certain iterations (epochs). Finally, the weights of the model get updated and analysis is done.

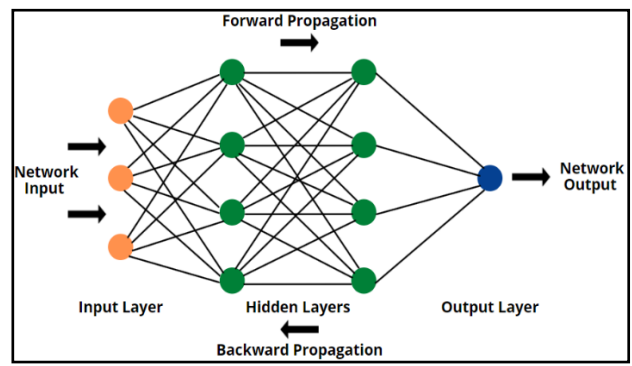

**Fig. 6: Artificial Neural Network Technique**

## 5. IMPLEMENTATION
The proposed approach for YouTube Ad view sentiment analysis with implementation is step-wise explained in this section. The steps involved are:

## 5.1 Dataset description
The dataset has been taken from Kaggle. It includes the metrics and a few other details of 15000 YouTube videos. The metric criteria consist of views, comments, likes, dislikes, duration, published date, and category. The attributes in the dataset are vivid, likes, views, adviews, dislikes, published, comments, category, and duration.

### 5.2 Importing the dataset and libraries
Initially, the pre-installed python libraries or packages like numpy, pandas, matplotlib, and seaborn were imported and used for cleaning data and visualization. Then the dataset in csv format was imported using pandas as a pandas dataframe. The number of features and samples in the data were explored.

### 5.3 Applying data visualization techniques
The Seaborn and matplotlib libraries were used for plotting. The individual features were plotted (as shown in fig. 7 and fig. 8) and the distribution of the data was analyzed. This was used to spot the outliers (if any) in the data which also helped the model to train better. The heatmap was also plotted (as shown in fig. 9) using the seaborn library which helped to visualize correlations with respect to each feature.

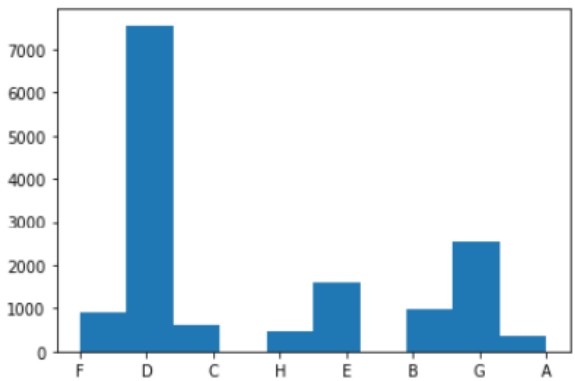

**Fig. 7: Individual plot of Category feature**

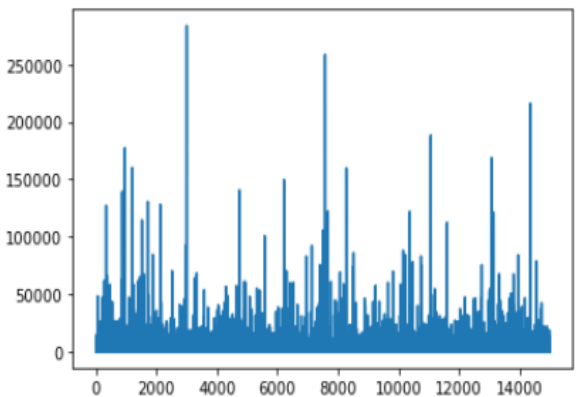

**Fig. 8: Individual plot of Likes feature**

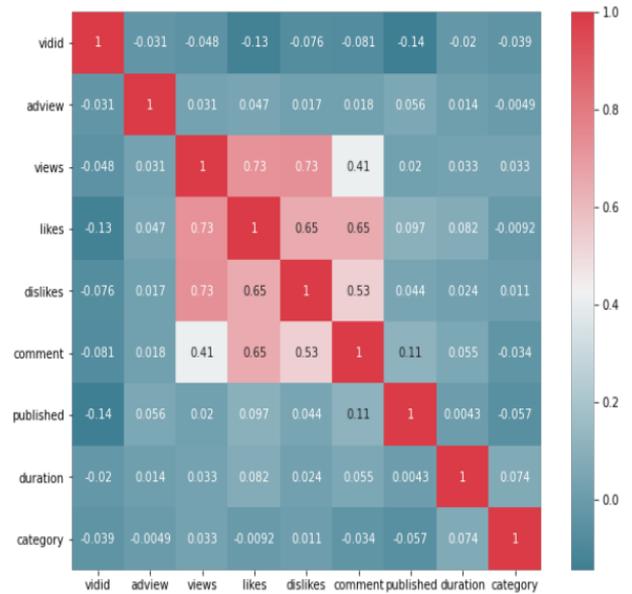

**Fig. 9: Heatmap of the data**

### 5.4 Data cleaning
Cleaning the dataset is one of the vital steps while interpreting and dealing with a machine learning problem. So, missing values from the data, which might affect the performance of the model, were removed.

### 5.5 Making necessary transformations
The categorical data and data which were in other formats, were converted into numerical form. The date, time, and label encoder functions were used for it. This process is also named as feature engineering.

### 5.6 Splitting and normalizing the data
The dataset was distinguished into the training and the testing data in the ratio 80:20 respectively. Then, normalization was done using MinMax Scaler (transforms variable in the range of 0 to 1), to verify if all the features were appropriately weighted in the training stage.

### 5.7 Training the model using classifiers
Several machine learning models like Linear Regression (LR), Support Vector Machine (SVM), Decision Tree (DT), and Random Forest (RF) were used to train the data. The scikit-learn library was used to import these models and train them, providing necessary labeled data or hyperparameters.

### 5.8 Training the model using ANN
Initially, the model architecture was defined including layers, number of neurons, activation function, and cost function. Then the model was trained for different epochs using keras, which resulted in the improvement of the model.

### 5.9 Analyzing the results
The results obtained were in the form of Root Mean Squared Error (RMSE). One machine learning model with minimum error and ANN model were selected for testing. Both the models were saved using keras and scikit-learn. Finally, the test data was used for the prediction of YouTube ad views from the chosen models.

## 6. RESULTS

The deep learning and machine learning models were applied to the dataset to perform experiments, in order to examine the performance of algorithms. Implementing all the algorithms resulted that, decision tree model obtained a minimum RMSE value. The results achieved from all the models are tabulated in Table 1.

**Table 1. Table captions should be placed above the table**

| Sr. No. | Techniques/Models | Root Mean Squared Error (RMSE) values |
|---|---|---|
| 1. | Linear Regression (LR) | 289.078 |
| 2. | Support Vector Machine (SVM) | 288.736 |
| 3. | Decision Tree (DT) | 260.193 |
| 4. | Random Forest (RF) | 264.935 |
| 5. | Artificial Neural Network (ANN) | 287.919 |

## 7. CONCLUSION

The paper proposes the approach for YouTube ad view prediction using advanced technologies like Deep Learning and Machine Learning. The techniques like Linear Regression (LR), Support Vector Machine (SVM), Decision Tree (DT), Random Forest (RF), and Artificial Neural Network (ANN) were used to train the model. The result achieved is that the decision tree algorithm obtained the minimum RMSE value of 260.193 and ANN acquired 287.919 RMSE value. As part of this research, an attempt has been made to study and implement various algorithms to enhance the feasibility of sentiment analysis where a massive amount of data can be analyzed without much time consumption. The aim is to analyze all the elements of sentiment analysis of most of the social media networks and platforms like Instagram, Twitter, etc., and build a system that will generate immediate results when the data is inserted into it. This study, hence, studies all the features and elements, implements them, and makes the prediction that is quite accurate. Therefore, this will overall contribute to the development of the advertising industry.